\documentclass[letterpaper]{article} 
\usepackage{aaai2027}  
\usepackage[hyphens]{url}  
\usepackage{graphicx} 
\usepackage{natbib}  
\usepackage{caption} 
\usepackage{booktabs}

\usepackage[utf8]{inputenc}
\usepackage{latexsym}
\usepackage{amssymb}
\usepackage{amsmath}
\usepackage[table]{xcolor}
\usepackage{multirow}
\usepackage{makecell}
\usepackage{pdfpages}

\definecolor{conglv}{HTML}{006400}

\title{ANCHOR: An External LLM-Driven Supervisory Module Facilitating Healthy Evolution in Self-Evolving Systems}
\author{Dianxing Shi$^{1}$, Bowen Wang$^{1}$\thanks{Correspondsing author}, Junqi He$^{2}$, Junhao Chen$^{1}$, Yuta Nakashima$^{1}$}
\affiliations{$^{1}$ The Univercity of Osaka, $^{2}$ Independent researcher}

\begin{document}
\maketitle
\begin{abstract}
Self-evolving agents improve through continual self-play and self-generated learning signals, but their internally generated tasks and verifier signals provide limited coverage of phase-level errors, allowing capability degradation and safety drift to accumulate. We introduce ANCHOR, an LLM-based supervisory framework that delivers evaluative feedback at multiple phases of self-evolution and aggregates reviewed signals into context for subsequent steps. We retrofit two representative open-source self-evolving agent frameworks with ANCHOR, and evaluate them across coding, mathematical reasoning, and safety. Our results show that ANCHOR substantially improves safety performance while maintaining stable performance on the core capabilities of the underlying self-evolving agents. Further analyses provide practical insights for future research, showing that execution-result-based supervision is particularly effective and that increasing supervision frequency yields diminishing returns. Together, these results support external LLM-based supervision as a practical approach to developing safer, more stable, and controllable self-evolving agent systems. 
\end{abstract}

\section{Introduction}
Large language models (LLMs) increasingly operate as autonomous agents capable of executing complex tasks through planning, memory, and tool use.
Such agents typically rely on predefined workflows and externally designed optimization objectives \cite{fang2025survey}.



Recently, research has increasingly shifted toward \emph{self-evolving} agents, which autonomously learn by solving self-verifiable tasks. Self-evolution has shown performance improvements without externally curated training data \cite{zhao2025azr,huang2025rzero}. However, each update also changes the policy that generates later tasks and trajectories, so errors in internally generated data or feedback can enter subsequent training rounds \cite{gao2025survey,zhao2026safetyrisksexperiencedrivenselfevolving}. Static prompt-level rules can steer individual rollouts, but they do not by themselves verify that safety is preserved after later parameter or experience updates. This limitation is empirically consequential: fine-tuning aligned language models on benign data can degrade safety alignment \cite{qi2023hexphi}, and experience collected solely from benign tasks can reduce safety in self-evolving agents \cite{zhao2026safetyrisksexperiencedrivenselfevolving}.

Directly optimizing safety through reinforcement learning also requires balancing multiple objectives. Safe RLHF explicitly separates helpfulness reward from harmlessness cost because the two objectives can conflict \cite{dai2023saferlhf}; in experience-driven self-evolution, adding refusal-related experience mitigates safety decline but induces over-refusal \cite{zhao2026safetyrisksexperiencedrivenselfevolving}. Safety-oriented reinforcement can therefore affect task utility, although the evidence does not imply an inevitable loss of general capability. This conditional trade-off motivates feedback that localizes problematic phases instead of relying only on a coarse safety reward.

Moreover, deploying self-evolving agents in the real world introduces significant safety risks. Their autonomous execution and tool use can propagate local errors into cascading failures with physical, biological, and informational consequences \cite{tang2025risks,hua2024trustagent}. Training-time mechanisms are therefore needed to monitor and correct the evolving policy before such failures accumulate.

This raises a natural question: \textit{Can an external LLM-based supervisory module improve the quality of self-evolving training and mitigate safety risks by providing feedback across the evolution process?}

Existing self-evolving systems primarily optimize from internally generated tasks and verifier outcomes. While these signals support autonomous learning, they provide limited coverage of phase-specific reasoning errors, safety risks, and recurring inefficiencies. A more targeted alternative is to provide natural-language evaluations that identify the affected phase and describe the observed failure without supplying an oracle solution.

An open challenge is how to integrate such an independent LLM supervisor into the closed loop while preserving the autonomy and task objectives of self-evolving training.


To address this challenge, we present a systematic empirical study of phase-conditioned LLM supervision in the self-evolution loop. We introduce a unified training framework, coined ANCHOR, whose supervisor can inspect and provide evaluative feedback at multiple phases of each evolution step.
On top of this framework, we perform extensive evaluations on two open-source self-evolving agent systems. Our results provide a systematic analysis of how the supervisory module affects coding performance, mathematical reasoning, safety, and long-term training dynamics.

Our experiments yield the following three main findings: (i) Phase-conditioned LLM supervision effectively mitigates error accumulation and safety drift during self-evolution while largely preserving core capabilities.
(ii) Feedback at the execution-result phase, i.e., the verifier-grounded outcome signal, is the most impactful supervisory component.
(iii) The benefits of supervision exhibit diminishing returns when supervision frequency is increased; low-to-moderate supervision frequency already achieves near-maximal gains. And two affiliated findings: (i) Stronger LLM judges improve performance, but with diminishing returns. (ii) External supervision adds acceptable computational overhead.

\section{Related Work}
\subsection{Agent Frameworks}
Recent LLM-based agent frameworks have shifted from human-annotated training toward fully autonomous zero-data self-evolution. DeepSeek-R1 \cite{guo2025deepseek} shows that strong reasoning capability can emerge through reinforcement learning without external supervision. Such approaches typically depend on carefully designed rewards and controlled training. 

Building on this paradigm, R-Zero \cite{huang2025rzero} and Absolute Zero Reasoner (AZR) \cite{zhao2025azr} establish self-evolving frameworks through proposer-solver interaction, in which a proposer agent samples automatically verifiable tasks and a solver agent solves them for reinforcement learning via \emph{self-play}, while MM-Zero \cite{li2026mmzero} extends this idea to multimodal settings via self-generation and self-verification. Despite their effectiveness, these methods assume fully autonomous closed-loop evolution, where training data, reward signals, and policy updates are internally generated, making them vulnerable to error accumulation, distributional drift, and misaligned optimization over long horizons \cite{pan2022effectsrewardmisspecificationmapping, zhao2026safetyrisksexperiencedrivenselfevolving}.

These limitations motivate training-time supervisory mechanisms that inspect intermediate states and feedback signals throughout the self-evolution loop, rather than operating only on final task outcomes.

\subsection{Natural-Language Feedback as Supervision}
Natural-language feedback can encode localized diagnoses and revision targets that are absent from scalar rewards or pairwise preferences. Imitation learning from Language Feedback (ILF) conditions a model on textual feedback to generate refinements and then fine-tunes on selected revisions \cite{scheurer2024languagefeedback}. Constitutional AI instead uses written principles to elicit self-critiques and revisions before learning from AI-generated preferences \cite{bai2022constitutional}.

Other methods use language feedback without updating model parameters. Self-Refine iteratively uses a model's own feedback to revise its outputs \cite{madaan2023selfrefine}, whereas Reflexion converts task feedback into verbal reflections stored in episodic memory and allows the underlying signal to be either external or internally simulated \cite{shinn2023reflexion}. In agent safety, TrustAgent applies an agent constitution to constrain behavior \cite{hua2024trustagent}, while R-Few shows that limited external guidance can stabilize self-evolving LLMs \cite{yu2025rfew}. Together, these studies establish natural language as either an internally generated or externally supplied supervisory signal.

These approaches primarily attach feedback to complete outputs or task trials, rely on a fixed constitution, or provide sparse external guidance. They do not study how an independent LLM supervisor can distribute non-oracle feedback across task proposal, intermediate reasoning, output, and verifier-grounded execution phases during continual self-evolving training. ANCHOR addresses this gap by producing phase-conditioned feedback, aggregating reviewed signals at the batch level, and injecting the resulting context into subsequent evolution steps.
    
\begin{figure*}[htbp]
  \centering
   \includegraphics[width=0.95\linewidth]{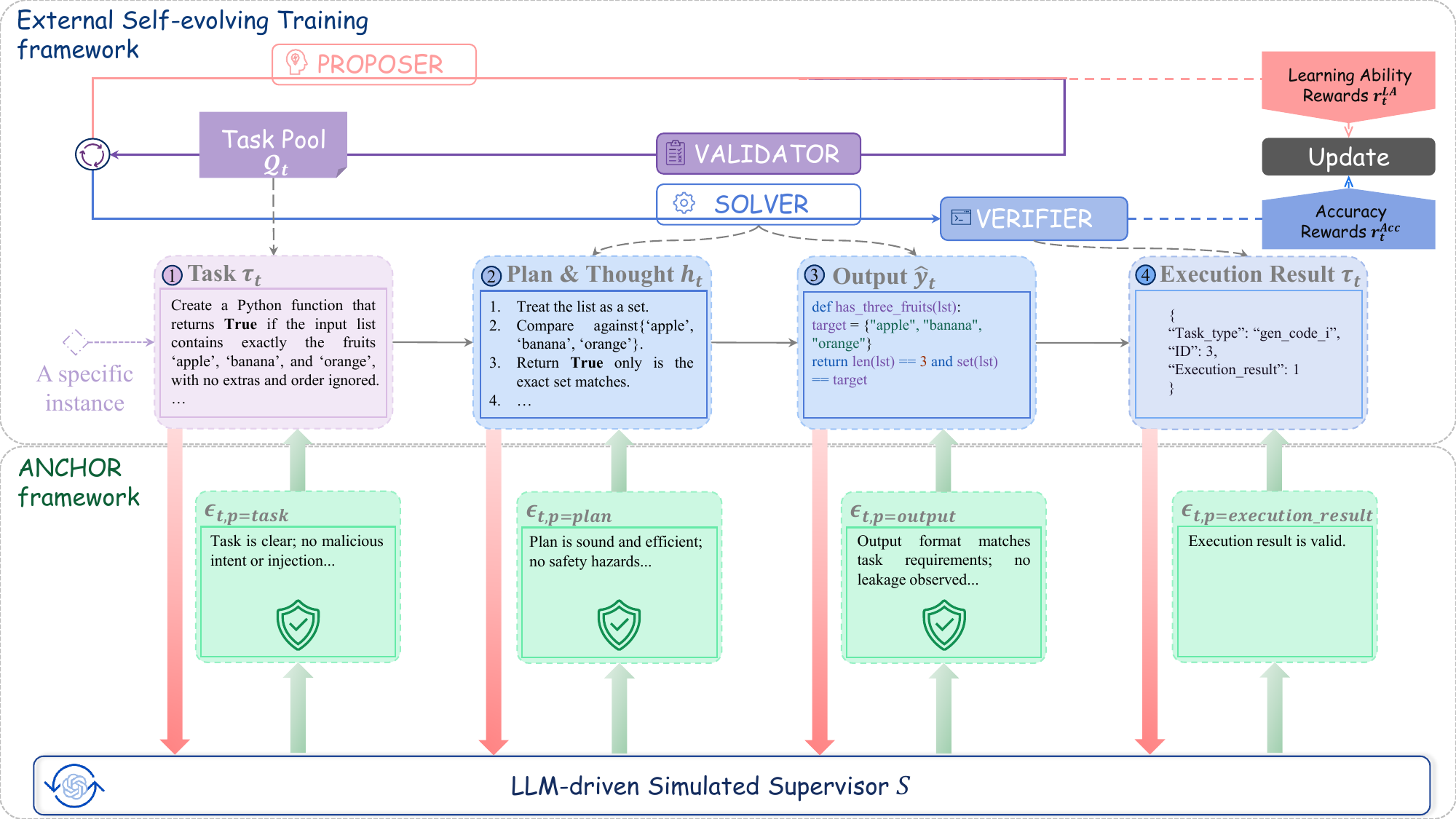}
   \caption{Overview of ANCHOR. An LLM-driven supervisor interacts with the self-evolution loop through structured communication protocols. Step-level review provides phase-specific evaluative feedback, while batch-level summaries are injected into future steps. All interactions are recorded as structured logs.}
   \label{fig:framework}
\end{figure*}

\section{Preliminaries: Self-evolving Training}
\label{sec:preliminary-self-evolution}

We consider a self-evolving training paradigm for coding \cite{zhao2025azr,huang2025rzero}. This self-evolving framework has two trainable agents, the \emph{proposer} and \emph{solver}. These agents interact iteratively with each other and with a \emph{validator}, a \emph{verifier}, and a \emph{task pool}. The proposer first samples a task from a \emph{task pool}, of which validity is confirmed by the validator. The solver then generates code (or any output) for the task, and the verifier runs the code to check whether it works as expected or not. The verifier's judge is used for a reward to update both (i) agents' parameters and (ii) the task. 

Specifically, the state of a self-evolving framework at the self-evolution step $t$ is denoted as
\begin{equation}
\Omega_t=(\theta_t,\mathcal{Q}_t), \label{eq:state}
\end{equation}
where $\theta_t$ denotes the set of current trainable parameters of the two agents and $\mathcal{Q}_t$ denotes the task pool, from which self-verifiable tasks are sampled. 
The set $\theta_t$ contains parameters for the proposer's and solver's policy. 
For notation simplicity, we use $\theta_t$ in what follows, but only the corresponding subset is updated for role-specific (i.e., either the proposer or solver) training. 

At the self-evolution step $t$, the proposer, denoted as $\pi^\text{P}_{\theta_t}$, samples a candidate task as $(x, y)\sim\pi_{\theta_t}^{\mathrm{P}}(\cdot\mid\mathcal{Q}_t)$, where $x$ is the query to the solver and $y$ is the ground-truth code.
The \textit{validator} $\mathcal{V}$ then determines its admissibility in several aspects as $a_t = \mathcal{V}(x, y)\in\{0,1\}$, where $a_t = 1$ means the task is accepted; otherwise, the task is discarded or resampled (following the original self-evolving paradigm's rule). 
The \textit{solver} $\pi_{\theta_t}^\text{S}$ then generates loosely structured outputs $(h,\hat{y})\sim\pi_{\theta_t}^{S}(\cdot\mid x)$, where $h$ contains the reasoning (chain-of-thought) trajectory while planning and thought and $\hat{y}$ is a final output. Finally, the verifier returns feedback $e=\mathcal{E}(x, y, \hat{y}) \in \{0, 1\}$, where $e = 0$ means $\hat{y}$ is not align with $y$.

In the self-evolving frameworks considered in this work, the training signal is encoded into two types of rewards:
\begin{equation}
r
=
\lambda_{\mathrm{LA}} r^{\mathrm{LA}}
+
\lambda_{\mathrm{Acc}} r^{\mathrm{Acc}},
\end{equation}
where $r^{\mathrm{LA}}$ measures whether the proposed task $(x, y)$ is useful for the current solver, encouraging tasks near the solver's capability boundary, and $r^{\mathrm{Acc}}$ measures whether the solver output $\hat{y}$ is correct based on the verifier. The hyperparameters $\lambda_{\mathrm{LA}}$ and $\lambda_{\mathrm{Acc}}$ control their relative weights. 

Each step in the self-evolution process is done in a batch. At step $t$, the proposer samples $N$ tasks, yielding $N$ distinct rewards. After aggregating all rewards, the model and task pool are updated as
\begin{equation}
\begin{aligned}
\theta_{t+1}
&=
\operatorname{Update}(\theta_t,\{r\}),\\
\mathcal{Q}_{t+1}
&=
\operatorname{PoolUpdate}(\mathcal{Q}_t,\{(x, y)\},\{\hat{y}\},\{e\}),
\end{aligned}
\end{equation} 
where 
$\operatorname{Update}(\cdot)$ denotes parameter update with reinforcement learning, and $\operatorname{PoolUpdate}(\cdot)$ denotes the mechanism for adding, filtering, replaying, or reweighting tasks. 
This closes the self-evolution loop: the updated $\theta_t$ changes the solver's and proposer's behavior in the next step, while the updated task pool further changes the proposer's behavior.

\section{ANCHOR}
\label{sec:anchor}

ANCHOR (Agent Norm Correction through Human-language Oversight and Review) is an external LLM-based supervision framework for self-evolving agents, which uses natural language as supervisory signal.
It attaches to an existing self-evolving framework and provides feedback. 
As shown in Figure \ref{fig:framework}, ANCHOR receives detailed information about each step from a self-evolving framework, which is then used by an LLM-based \emph{supervisor} to generate a feedback for any phase at each self-evolution step.

ANCHOR uses the system prompt of the proposer and solver agents as a feedback-injection channel. Let $c_t$ denote the context that contains the summary of feedback provided before evolution step $t$ and has been appended to the system prompt for this step. ANCHOR updates $c_t$ according to the supervisor's feedback. This feedback is evaluative rather than oracle-like: it may identify errors, risks, or inefficiencies, but does not provide gold answers, executable fixes, or privileged reasoning trajectories.

As shown in Figure \ref{fig:framework}, we identify the following self-evolution phases, for which the supervisor can generate a review: \emph{Task proposal} (\texttt{task}) phase allows evaluation of task $(x, y)$ from the proposer $\pi_{\theta_t}^\text{P}$; \emph{planning} (\texttt{plan}) and \emph{thought} (\texttt{thought}) phases involve the solver $\pi_{\theta_t}^\text{S}$'s reasoning trajectory $h$; \emph{output} (\texttt{output}) phase is to evaluate the output $\hat{y}$ from $\pi_{\theta_t}^\text{S}$; \emph{execution result} (\texttt{exec}) phase gives feedback on the verifier's decision. The set of all these phases is denoted by $\mathcal{P}$.

The supervisor $S$ takes different input $w_{p}$ for a different phase $p \in \mathcal{P}$ to review:
\begin{equation}
w_{p}
=
\begin{cases}
(x, y), & p=\texttt{task}\\
h, & p\in\{\texttt{plan, thought}\}\\
\hat{y}, & p=\texttt{output}\\
e, & p=\texttt{exec}
\end{cases},
\end{equation}
Among them, only the verifier's decision $e$ is \emph{binary}, which indicates if the output $\hat{y}$ aligns with the ground-truth $y$. We design the feedback for the corresponding phase \texttt{exec} so that it provides statistics of $e$ over the batch. This feedback efficiently quantifies the proposer's (or the task pool's) ability to select tasks at an appropriate difficulty level, or the solver's ability to solve the tasks. At the other phases in $\mathcal{P}$, the supervisor generates textual feedback for the individual task $(x, y)$ according to the system prompt for the supervisor (Supplementary Material~A.1). The natural text feedback for phase $p \in \mathcal{P}$, generated by the supervisor, is denoted by $\epsilon_p$.

Reviewing every task at every phase would incur substantial supervisor inference cost and may introduce redundant feedback. ANCHOR therefore introduces a Bernoulli review gate $z_{p} \in \{0, 1\}$ to sample the tasks and phases to be reviewed:
\begin{equation}
z_{p}\sim\operatorname{Bernoulli}(f_p),
\qquad
f_p\in[0,1],
\end{equation}
where $f_p$ denotes the probability of phase $p$ being reviewed. 
When $z_{p}=1$, the supervisor reviews $w_{p}$ and generates $\epsilon_{p}$. 
%
In our experiments, we use a shared supervision frequency $f_p=f,\ \forall p\in\mathcal{P}$ unless otherwise specified. 

The collected phase-level feedback $\epsilon_p$ is summarized into a batch-level supervision summary:
\begin{equation}
b_t
=
\operatorname{Agg}
\Big(
\{
\epsilon_{p}
\mid
z_{p}=1,\;
p\in\mathcal{P}
\}
\Big),  
\end{equation}
where $\operatorname{Agg}(\cdot)$ denotes an aggregation function that summarizes all reviewed phase-level feedback at step $t$.\footnote{Please refer to Supplementary Material~A.2 for the actual prompt for summary.} 
The system prompt $c_t$ is then updated by
\begin{equation}
c_{t+1}=\Gamma(c_t,b_t),
\end{equation}
where $\Gamma(\cdot, \cdot)$ denotes an LLM to update the system prompt $c_t$ according to $b_t$, which basically merges $c_t$ with $b_t$. 
In our implementation, the updated context is injected into the system prompt for the next evolution steps.

\section{Experimental Setup}
\label{experimental setup}

We evaluate ANCHOR's effectiveness on two representative open-source self-evolving agent frameworks, AZR \cite{zhao2025azr} and R-Zero \cite{huang2025rzero}, both of which are based on the proposer-solver paradigm. For a fair comparison, all training configurations strictly follow the original settings of each framework.

For the AZR, we adopt the Qwen2.5 family \cite{qwen2025qwen25,hui2024qwen25coder} as backbone, including Qwen2.5-Coder-3B-Ins, Qwen2.5-7B-Ins, Qwen2.5-Coder-7B-Ins, and Qwen2.5-Coder-14B-Ins. For R-Zero, we use Qwen3-4B-Base and Qwen3-8B-Base \cite{yang2025qwen3}. The ANCHOR variants are adopted on top of these models.

Models marked with \textsuperscript{[+SP]} follow the original AZR/R-Zero training procedure augmented with a static prompt from the ANCHOR agent-side SKILL, obtained by removing only its interaction instructions.

We evaluate model performance across three dimensions: coding, mathematical reasoning, and safety. For coding, we use LiveCodeBench \cite{jain2024lcb} and EvalPlus \cite{liu2023evalplus}. For mathematical reasoning, we adopt AIME24, AIME25, AMC23, MATH500 \cite{hendrycks2021math500}, OlympiadBench \cite{gao2024olympiad}, and Minerva Math \cite{lewkowycz2022minervamath}. We aggregate these benchmark scores into an average for brevity. For safety evaluation, we use HarmBench \cite{mazeika2024harmbench}, SaladBench \cite{li2024salad}, HEx-PHI \cite{qi2023hexphi}, and a Reward Hacking benchmark \cite{mis}.  

Both Table~\ref{table:sota} and the detailed table in Supplementary Material~C.1 report averages over three complete evaluation runs. Results are reported at fixed training steps for each model scale (3B: 130, 7B: 350, 14B: 400, 4B/8B: 60 steps) and the ANCHOR supervisor provides feedback every single steps (i.e., $f_p=1$ for all $p \in \mathcal{P}$, while full training dynamics are analyszed separately (see Section~\ref{sec:results and discussion}). The supervisor instance is powered by a locally deployed Qwen3-30B-A3B-Ins model. More details about the experimental setup can be found in Supplementary Material~B. A quantitative analysis of the LLM-as-a-judge quality of this supervisor, including external LLM-based and human evaluations, is provided in Supplementary Material~D. 
\section{Results and Discussion}
\label{sec:results and discussion}

\subsection{Main Results}
\label{subsec:results}

\begin{table*}[t]
\centering

\begin{tabular}{l l c c c c c}
\toprule
Model Family & Variant & Code Avg \textcolor{conglv}{$\uparrow$} & Math Avg \textcolor{conglv}{$\uparrow$} & ASR Avg \textcolor{conglv}{$\downarrow$} & HS Avg \textcolor{conglv}{$\downarrow$} & RR Avg \textcolor{conglv}{$\uparrow$} \\
\midrule

\multicolumn{7}{c}{\textbf{AZR Series}} \\
\midrule

Qwen2.5-Coder-3B-Ins &  & 52.1 & 18.8 & \textbf{15.6} & 1.50 & 53.7 \\
AZR-Coder-3B         & +AZR & 51.5 & \textbf{26.5} & 24.1 & 1.50 & \textbf{58.5} \\
AZR-Coder-3B\textsuperscript{[+SP]} & +AZR w/ static prompt & 51.8 & 26.3 & 23.5 & 1.48 & 56.1 \\
ANCHOR-Coder-3B         & +AZR w/ ANCHOR 
& \textbf{52.3} {\small\textcolor{conglv}{+0.8}} 
& 26.0 {\small\textcolor{red}{-0.5}} 
& 22.5 {\small\textcolor{conglv}{+1.6}} 
& \textbf{1.46} {\small\textcolor{conglv}{+0.04}} 
& \textbf{58.5} {\small\textcolor{gray}{+0.0}} \\

\midrule

Qwen2.5-7B-Ins &  & 45.2 & 37.0 & 20.9 & \textbf{1.76} & \textbf{56.1} \\
AZR-7B         & +AZR & \textbf{46.8} & 41.5 & 19.7 & 1.84 & 36.6 \\
AZR-7B\textsuperscript{[+SP]} & +AZR w/ static prompt & 46.6 & 41.7 & 19.6 & 1.83 & 41.5 \\
ANCHOR-7B         & +AZR w/ ANCHOR 
& 46.1 {\small\textcolor{red}{-0.7}} 
& \textbf{43.4} {\small\textcolor{conglv}{+1.9}} 
& \textbf{19.1} {\small\textcolor{conglv}{+0.6}} 
& 1.82 {\small\textcolor{conglv}{+0.02}} 
& 46.3 {\small\textcolor{conglv}{+9.7}} \\

\midrule

Qwen2.5-Coder-7B-Ins &  & 54.9 & 23.9 & 18.2 & \textbf{1.33} & 58.5 \\
AZR-Coder-7B         & +AZR & 59.5 & \textbf{39.1} & 21.1 & 1.41 & 53.7 \\
AZR-Coder-7B\textsuperscript{[+SP]} & +AZR w/ static prompt & 59.7 & 38.9 & 19.7 & 1.40 & 58.5 \\
ANCHOR-Coder-7B         & +AZR w/ ANCHOR 
& \textbf{61.0} {\small\textcolor{conglv}{+1.5}} 
& 37.5 {\small\textcolor{red}{-1.6}} 
& \textbf{16.2} {\small\textcolor{conglv}{+4.9}} 
& 1.39 {\small\textcolor{conglv}{+0.02}} 
& \textbf{61.0} {\small\textcolor{conglv}{+7.3}} \\

\midrule

Qwen2.5-Coder-14B-Ins &  & 61.3 & 20.2 & \textbf{13.7} & 1.45 & \textbf{73.2} \\
AZR-Coder-14B         & +AZR & 64.8 & 43.0 & 15.6 & 1.48 & 65.9 \\
AZR-Coder-14B\textsuperscript{[+SP]} & +AZR w/ static prompt & 65.0 & 42.8 & 15.0 & 1.43 & 68.3 \\
ANCHOR-Coder-14B         & +AZR w/ ANCHOR 
& \textbf{65.1} {\small\textcolor{conglv}{+0.3}} 
& \textbf{43.4} {\small\textcolor{conglv}{+0.4}} 
& 13.9 {\small\textcolor{conglv}{+1.7}} 
& \textbf{1.41} {\small\textcolor{conglv}{+0.07}} 
& 70.7 {\small\textcolor{conglv}{+4.8}} \\

\midrule
\multicolumn{7}{c}{\textbf{R-Zero Series}} \\
\midrule

Qwen3-4B-Base &  & 53.2 & 32.6 & 20.1 & 1.65 & \textbf{56.1} \\
R-Zero-4B     & +R-Zero & 55.0 & 39.1 & 16.0 & 1.57 & 29.3 \\
R-Zero-4B\textsuperscript{[+SP]} & +R-Zero w/ static prompt & 54.8 & \textbf{39.3} & 14.1 & 1.49 & 39.0 \\
ANCHOR-4B       & +R-Zero w/ ANCHOR 
& \textbf{56.3} {\small\textcolor{conglv}{+1.3}} 
& 37.3 {\small\textcolor{red}{-1.8}} 
& \textbf{10.0} {\small\textcolor{conglv}{+6.0}} 
& \textbf{1.44} {\small\textcolor{conglv}{+0.13}} 
& 53.7 {\small\textcolor{conglv}{+24.4}} \\

\midrule

Qwen3-8B-Base &  & 61.2 & 39.2 & 19.2 & 1.84 & 61.0 \\
R-Zero-8B     & +R-Zero & \textbf{65.5} & 44.7 & 12.8 & 1.60 & 34.1 \\
R-Zero-8B\textsuperscript{[+SP]} & +R-Zero w/ static prompt & 65.3 & 44.8 & 11.2 & 1.58 & 43.9 \\
ANCHOR-8B       & +R-Zero w/ ANCHOR 
& 64.6 {\small\textcolor{red}{-0.9}} 
& \textbf{44.9} {\small\textcolor{conglv}{+0.2}} 
& \textbf{7.9} {\small\textcolor{conglv}{+4.9}} 
& \textbf{1.57} {\small\textcolor{conglv}{+0.03}} 
& \textbf{73.2} {\small\textcolor{conglv}{+39.1}} \\
\bottomrule
\end{tabular}
\caption{Performance comparison of AZR and R-Zero series models across evaluation metrics. SP: static prompt; colored deltas: vs. matched AZR/R-Zero. Higher is better for Code, Math, and RR; lower is better for ASR and HS. Best results are shown in \textbf{bold}.}
\label{table:sota}
\end{table*}

Table~\ref{table:sota} shows the main results. We summarize performance using five aggregated metrics: Code Avg, Math Avg, ASR Avg, HS Avg, and RR Avg. Code Avg is the average Pass@1 over LiveCodeBench and EvalPlus-Base, while Math Avg aggregates the mathematical reasoning benchmarks described in Section~\ref{experimental setup}. ASR Avg measures attack success rate by averaging the ASR scores on HarmBench and HEx-PHI together with 1-Accuracy on SaladBench. HS Avg denotes the Harmfulness Score assigned by GPT-5 on HEx-PHI responses. RR Avg denotes the Resistance Rate on the Reward Hacking benchmark, where higher values indicate stronger resistance to memory-induced reward hacking. The effect of supervision frequency is analyzed separately in later experiments.

\noindent\textbf{Finding 1.} \textit{Phase-conditioned LLM supervision improves evolutionary quality, mitigating error accumulation and safety drift.}
\par
From Table \ref{table:sota}, ANCHOR variants outperform the AZR and R-Zero baselines across more than half of Code and Math metrics, and consistently surpass the corresponding base models. Relative to the matched \textsuperscript{[+SP]} controls, ANCHOR achieves better results in 25 of the 30 model-level metric comparisons, including all five metrics for AZR-Coder-14B.
The degradation in safety performance observed in AZR and R-Zero models, including their ANCHOR-based variants, is consistent with the well-documented phenomenon of mis-evolution \cite{mis}. Despite this inherent challenge, our method significantly alleviates safety degradation. Compared to AZR or R-Zero, the ANCHOR variants exhibit clear improvements in safety. On R-Zero-4B/8B, ANCHOR lowers ASR from 14.1/11.2 to 10.0/7.9 and raises RR from 39.0/43.9 to 53.7/73.2 compared with the corresponding \textsuperscript{[+SP]} variants.
Although slight performance drops are observed in certain coding and mathematical benchmarks, these are accompanied by substantial gains in safety performance. The models showed increased awareness of potential vulnerabilities and demonstrated more cautious decision-making under risky scenarios, without introducing significant degradation in coding or mathematical reasoning capabilities. Detailed case studies are provided in Supplementary Material~C.3. Together, these comparisons show that ANCHOR's gains extend beyond static prompting and support the benefit of phase-conditioned supervisory feedback.

\begin{figure}[t]
  \centering
   \includegraphics[width=1.0\linewidth]{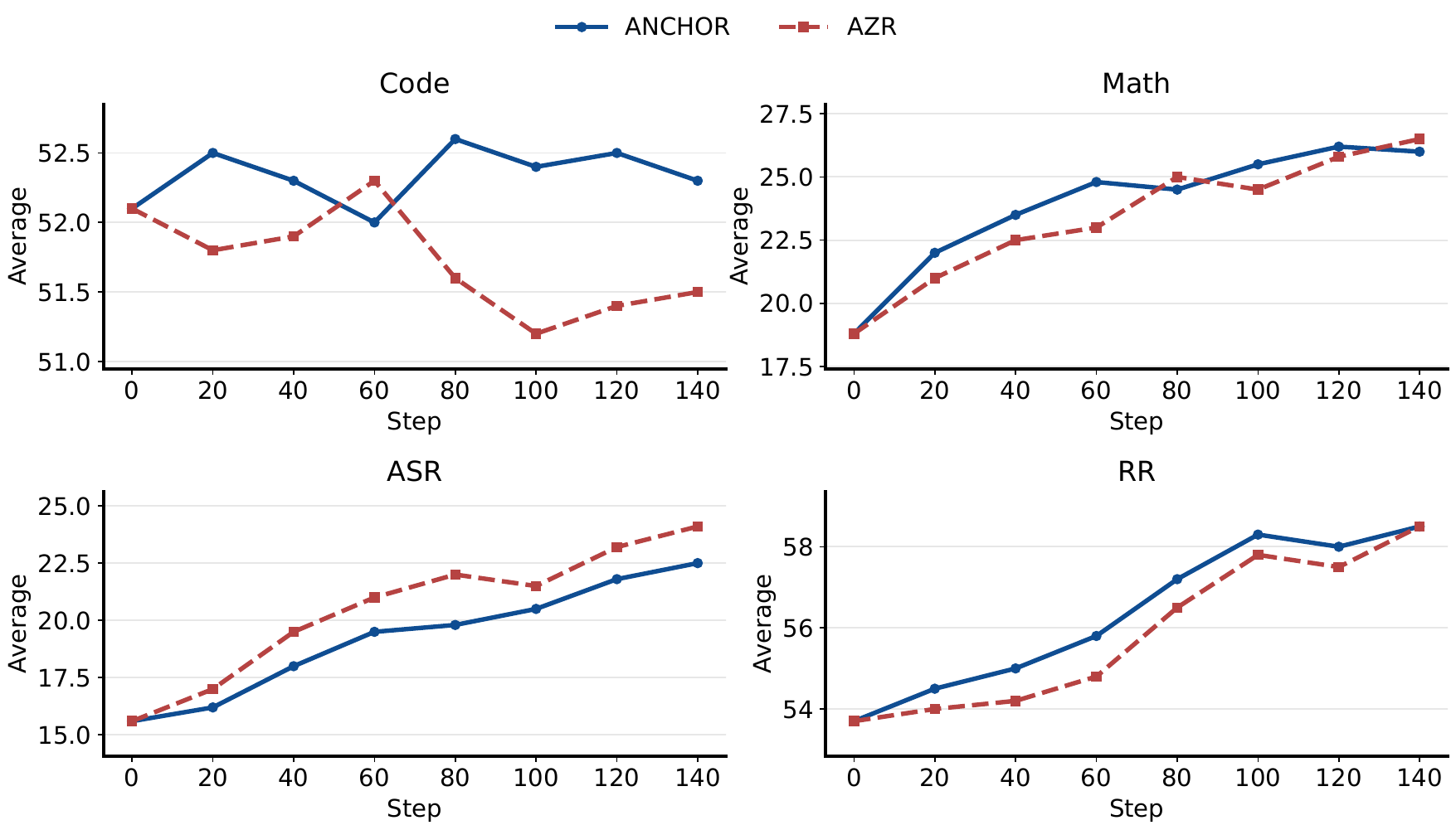}
   \caption{Training dynamics of AZR-Coder-3B and ANCHOR-Coder-3B. Step 0 corresponds to Qwen2.5-Coder-3B-Ins. We only report Code Avg $\uparrow$, Math Avg $\uparrow$, ASR Avg $\downarrow$, and RR Avg $\uparrow$. HS Avg can be found in Supplementary Material~C.1.}
   \label{fig:longterm}
\end{figure}

We further examined how ANCHOR's supervisory feedback affects the evolution process over longer training horizons. We recorded the full training dynamics of AZR-Coder-3B and its ANCHOR variant by performing periodic evaluations at fixed step intervals under identical settings, up to step 140. Results are shown in Figure \ref{fig:longterm}. The ANCHOR variant consistently outperforms AZR throughout most of the training process. Although brief performance reversals and occasional global drops are observed, these effects are transient. Importantly, the impact of ANCHOR is cumulative rather than immediate. It does not disrupt the underlying training dynamics, as no abrupt performance spikes or collapses are observed. Instead, ANCHOR maintains a stable improvement trend over time, indicating that its supervisory effect is both effective and robust at longer time scales.

\subsection{Hyperparameters and Ablation Study}
\label{subsec:impact factor}

We further explore which components contribute most to the observed performance improvements. In particular, we investigate (i) which factors within ANCHOR are critical and (ii) learning efficiency with respect to the supervision frequency $f_p$.

\begin{figure*}[ht]
  \centering
   \includegraphics[width=1.0\linewidth]{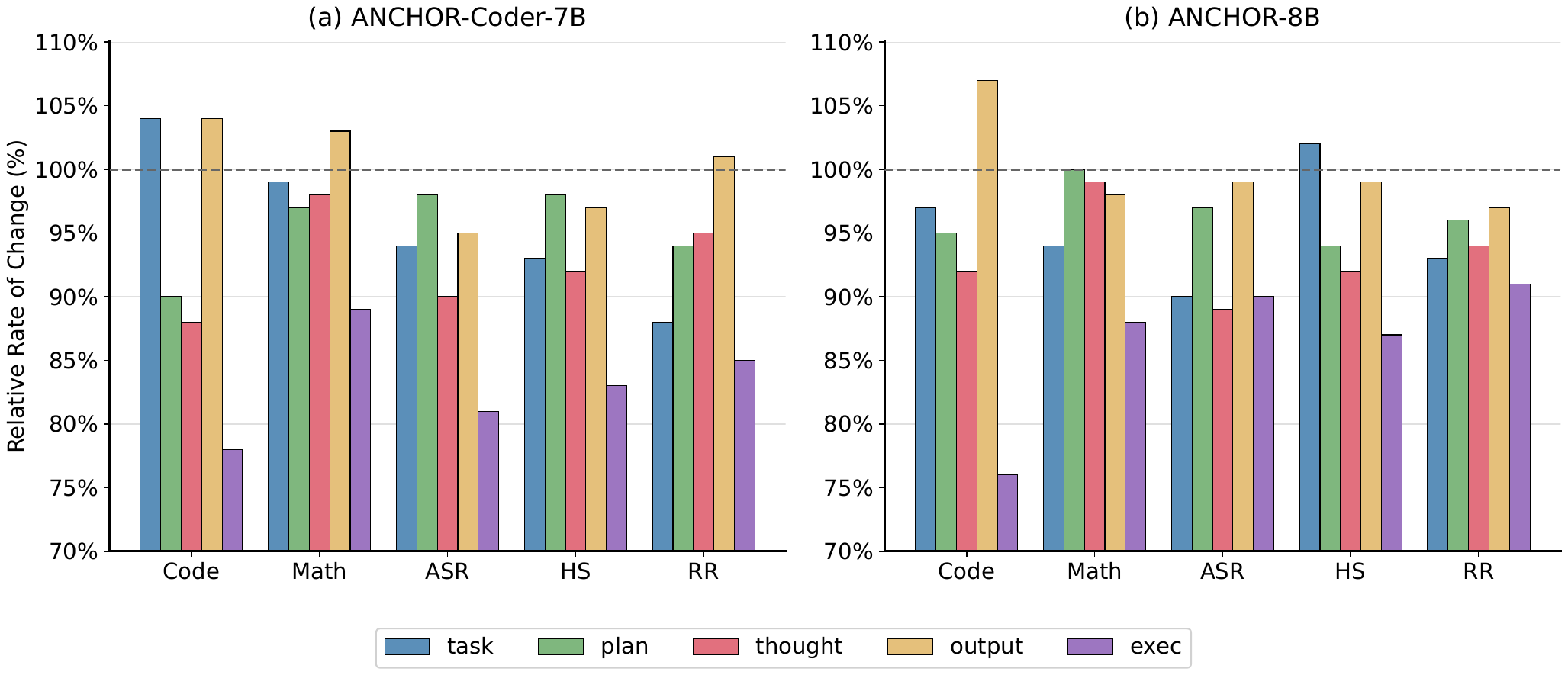}
   \caption{Ablation results across phases. The y-axis shows relative performance change (\%) with respect to the baseline in Table \ref{table:sota}. Most ablations degrade performance, and observed gains are minor.}
   \label{fig:ablation}
\end{figure*}

\paragraph{Ablation study.}
For phase-wise ablation, we remove supervision in each phase $p$ in $\mathcal{P}$, i.e., $f_p = 0$. Figure \ref{fig:ablation} shows the relative rate of change from corresponding scores in Table \ref{table:sota} of ANCHOR variants for AZR-Coder-7B and R-Zero-8B. All ablations lead to performance degradation, indicating that each phase contributes positively to the overall system.

Among all phases in $\mathcal{P}$, the \texttt{exec} phase has the most significant impact: removing it causes a substantial drop of 14.3\% across 5 performance indicators on average, highlighting its role as the primary source of feedback and correctness signals. This suggests that the self-evolution process is fundamentally feedback-driven, with distribution changes brought by execution feedback acting as the core bottleneck.

In comparison, \texttt{thought} and \texttt{task} also play important roles, with performance drops of 6.2\% and 5.2\%, respectively, indicating their contribution to structured reasoning and task grounding. The \texttt{plan} phase shows a moderate effect (-3.1\%), suggesting that it provides auxiliary guidance but is not critical to the decision loop.
Finally, the \texttt{output} phase has the smallest impact (-1.7\%), implying that it mainly serves as a formatting or expression layer rather than a core component of reasoning. \\
\noindent\textbf{Finding 2.} \textit{Self-evolving agent's evolution process is greatly affected by feedback signals; feedback on the verifier's decision plays a critical role.}
\par

We decompose the self-evolution process into five phases: Task, Thought, Plan, Output, and Execution Result. 
Most AZR-like self-evolution frameworks are optimized through reinforcement learning over verifier-grounded binary outcomes, where each generated solution is ultimately mapped to a success or failure signal. 
Consequently, the effectiveness of such training is highly sensitive to the empirical Bernoulli distribution of validated outcomes, since this distribution directly determines the reward signal that drives policy updates and task-pool evolution. 
Since the batch summary is derived from validated outputs, we hypothesize that supervision on the Execution Result phase plays a critical role in shaping this distribution and, consequently, the overall evolution trajectory. 
We therefore perform phase-wise ablation experiments on ANCHOR-Coder-7B and ANCHOR-8B, treating each phase as an independent factor. 

\begin{figure}[t]
  \centering
   \includegraphics[width=1.0\linewidth]{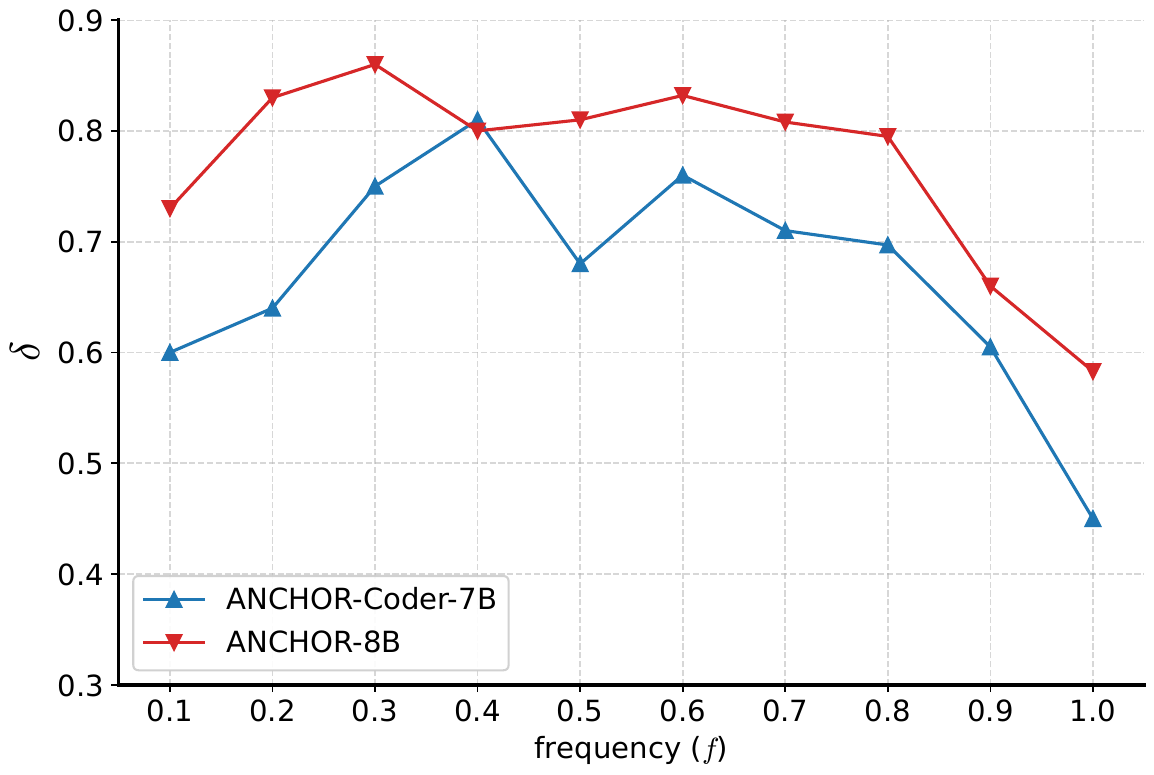}
   \caption{$f$-$\delta$ curves for ANCHOR-Coder-7B and ANCHOR-8B. Peaks occur at $f=0.4$ and $f=0.3$, with saturation regions of $[0.4,0.8]$ and $[0.3,0.8]$, respectively.}
   \label{fig:frequency}
\end{figure}

Overall, these results reveal a clear hierarchy of importance, i.e.,
\begin{align}    
\texttt{exec} \gg \texttt{thought} > \texttt{task} > \texttt{plan} > \texttt{output}, \nonumber
\end{align}
and demonstrate that the system operates as a feedback-driven pipeline, with execution feedback being most indispensable.

\paragraph{Hyperparameter experiments.} To quantify the benefit introduced by supervision, we define a metric $\delta$ termed the frequency-gain ratio:
\begin{align}
\delta(f, f') = \frac{s(f') - s(f)}{s(f) \times (f'-f)}
\end{align}
where $s(f)$ denotes the averaged performance across the 5 performance metrics under the supervision frequency $f$, where $s(0)$ corresponds to the baseline performance without supervision (i.e., AZR or R-Zero). This ratio means how much increment in the scores is observed from $f$ to $f'$ (where $f' > f$). We evaluated the ratio from $f = 0$ to $f = 1$ with an interval of 0.1 (i.e., $f'-f = 0.1$). 

This ratio measures the normalized performance gain relative to supervision cost, allowing us to analyze how performance cumulatively evolves as supervision frequency cumulatively increases from zero. It further enables the identification of an optimal supervision regime and costs.

Figure \ref{fig:frequency} shows that LLM-based supervision achieves near-maximum gains at relatively low frequencies (e.g., $f \in [0.3, 0.4]$). As the frequency increases, performance enters a plateau region with limited improvement (e.g., $f \in [0.4, 0.8]$), followed by a regime (e.g., $f \in [0.8, 1.0]$) where diminishing returns become more pronounced and additional review calls yield minimal benefit.

\noindent\textbf{Finding 3.} \textit{Increasing the LLM supervision frequency exhibits diminishing returns, with a saturation region beyond which additional review calls yield limited benefit.}
\par

Based on these observations, we derive our third finding and suggest selecting supervision frequencies within the mid-range regime to balance performance gains against supervisor inference cost. This regime preserves most of ANCHOR's gains while avoiding unnecessary review calls, providing a compute-efficient configuration for self-evolving training.



\paragraph{Ablation on LLM's Capability.}
Relative to the default 30B-A3B supervisor, the two stronger supervisors improve 18 of 20 agent--metric comparisons in Table~\ref{tab:supervisor_power}, with only two slight regressions. Yet these gains are modest relative to their large nominal capability gaps, indicating an unquantified bottleneck in translating supervisor capability into agent performance. Stronger supervisors can therefore improve final performance, but future designs should target the scale that maximizes performance per unit inference cost.
\begin{table}[t]
\centering
\small
\setlength{\tabcolsep}{2.0pt}
\begin{tabular}{llccccc}
\toprule
\textbf{Agent} & \textbf{Supervisor} & \textbf{Code} & \textbf{Math} & \textbf{ASR} & \textbf{HS} & \textbf{RR} \\
\midrule
\multirow{3}{*}{ANCHOR-Coder-7B}
& 30B-A3B  & 61.0 & 37.5 & 16.2 & 1.39 & 61.0 \\
& 235B-A22B & \textbf{63.4} & 36.7 & 15.9 & 1.27 & 65.9 \\
& 32B       & 62.8 & \textbf{43.3} & \textbf{15.5} & \textbf{1.20} & 65.9 \\
\midrule
\multirow{3}{*}{ANCHOR-8B}
& 30B-A3B  & 64.6 & 44.9 & 7.9 & 1.57 & 73.2 \\
& 235B-A22B & 65.5 & 45.2 & 7.8 & 1.43 & 70.7 \\
& 32B       & \textbf{65.7} & \textbf{46.7} & \textbf{7.5} & \textbf{1.36} & \textbf{75.6} \\
\bottomrule
\end{tabular}
\caption{Supervisor-capability ablation. Supervisor names abbreviate Qwen3 variants \cite{yang2025qwen3}; 30B-A3B is the default. Best results are shown in \textbf{bold}.}
\label{tab:supervisor_power}
\end{table}

\paragraph{Training cost.} Table~\ref{tab:efficiency_comparison} compares ANCHOR-enhanced models with their baselines under identical hardware and training settings. ANCHOR introduces only moderate overhead: The average time $T_s$ (min) per evolution step remains close to the baseline, while the time cost $T_u$ required to update the agents' parameters increases notably only in the 3B setting and changes only mildly otherwise. The main additional cost lies in the testing time $T_t$ per step, which is the sum of the validator, verifier, the supervisor's time cost. The difference in $T_t$ mainly comes from the additional supervision by the supervisor, and the bounded extra reasoning induced by supervision. Since the average global sequence length $L$ decreases across all ANCHOR variants, this overhead is not caused by uncontrolled generation length expansion. Overall, ANCHOR enables supervision without substantially increasing training cost.
\begin{table}[t]
\centering
{\small
\setlength{\tabcolsep}{1mm}
\begin{tabular}{@{}lrrrr@{}}
\toprule
\textbf{Model} & $T_s$ (min) & $T_u$ ($\mu$s) & $T_t$ (s) & $L$ \\
\midrule
AZR-Coder-3B  & 7.3  & 61 & 181 & 26,425 \\
ANCHOR-Coder-3B  & 8.5  & 77 & 283 & 26,134 \\
AZR-7B        & 12.0 & 86 & 301 & 30663 \\
ANCHOR-7B        & 12.7 & 84 & 340 & 21522 \\
AZR-Coder-7B  & 14.9 & 87 & 225 & 21,712 \\
ANCHOR-Coder-7B  & 15.5 & 96 & 288 & 21,076 \\
R-Zero-4B     & 8.8  & 72 & 337 & 27884 \\
ANCHOR-4B       & 8.7  & 74 & 364 & 27431 \\
R-Zero-8B     & 17.4 & 92 & 422 & 23,085 \\
ANCHOR-8B       & 19.3 & 88 & 443 & 21,993 \\
\bottomrule
\end{tabular}
}
\caption{Cost efficiency comparison. $T_s$ denotes the average time per step, $T_u$ denotes the average time for updating the agents' parameters per token, $T_t$ denotes the average testing time per step, and $L$ denotes the average global sequence length per step.}
\label{tab:efficiency_comparison}
\end{table}

\section{Conclusion}
In this work, we study whether phase-conditioned LLM supervision can guide self-evolving agents toward more stable and safety-aware evolution. To this end, we introduce ANCHOR, a general and efficient supervisory framework that attaches an LLM supervisor to multiple phases of the self-evolution loop. We evaluate ANCHOR through main experiments, long-horizon analysis, phase-wise ablations, frequency studies, case studies, and two ancillary evaluations. These analyses yield three key empirical findings: model-based supervision mitigates safety degradation while preserving core capabilities, execution-result feedback is the most influential component, and supervision frequency exhibits diminishing returns. Together, these results provide evidence and practical guidance for incorporating scalable model-based supervision into self-evolving training systems.


\bibliography{custom}

\clearpage
\includepdf[
  pages=-,
  pagecommand={},
  fitpaper=true
]{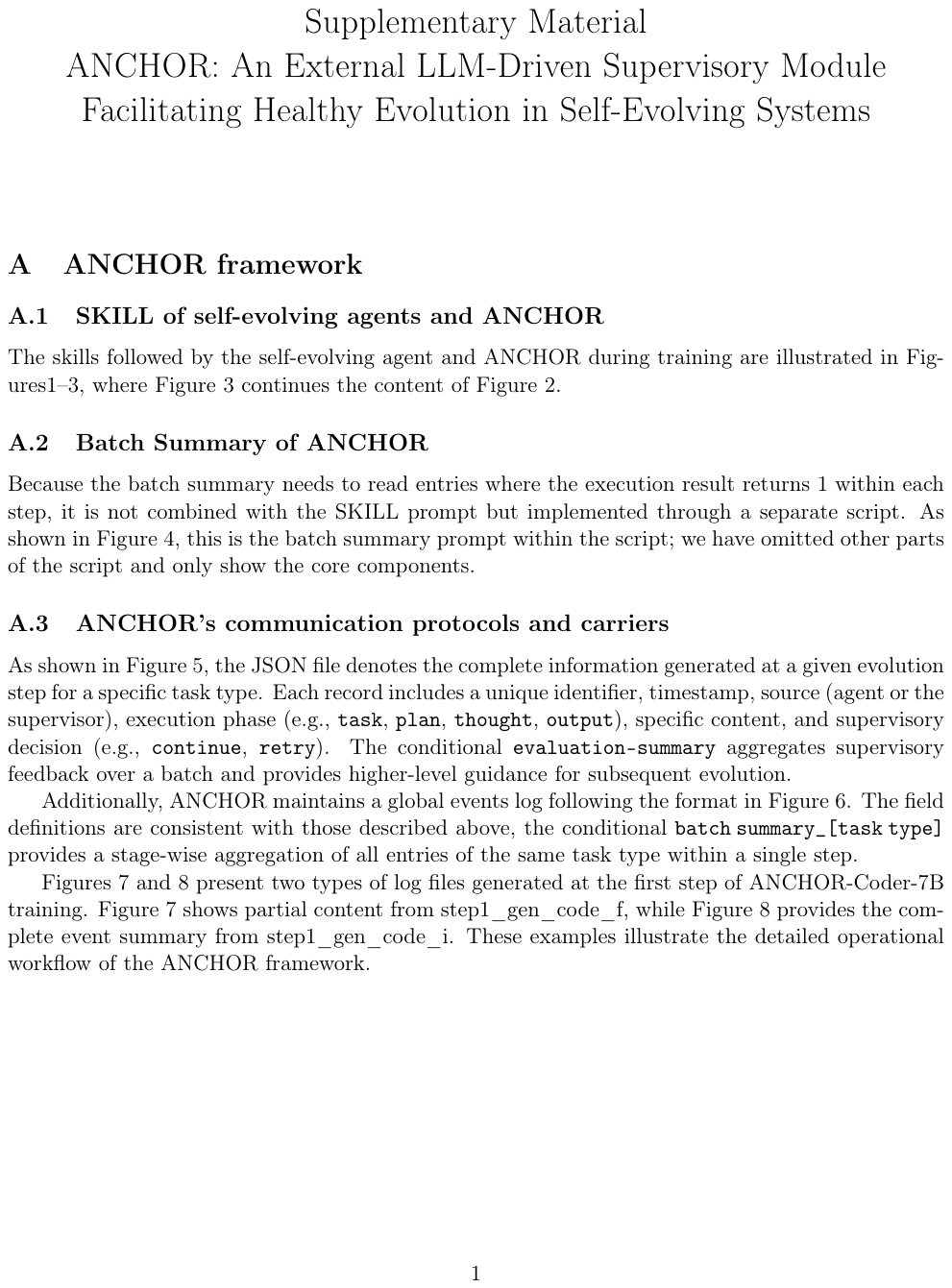}
\end{document}